\documentclass[10pt,twocolumn,letterpaper]{article}

\usepackage[pagenumbers]{cvpr} 

\usepackage[dvipsnames]{xcolor}

\definecolor{cvprblue}{rgb}{0.21,0.49,0.74}
\usepackage[pagebackref,breaklinks,colorlinks,citecolor=cvprblue]{hyperref}

\usepackage{xcolor}    
\usepackage{amsmath}
\usepackage{amssymb}
\usepackage{graphicx}
\usepackage{enumitem}
\usepackage{algorithm}
\usepackage{wrapfig}
\usepackage{algpseudocode}
\usepackage{bm}
\usepackage{float}
\usepackage{caption}
\usepackage{multirow}

\def\paperID{7685} 
\def\confName{CVPR}
\def\confYear{2024}

\title{TextCraftor: Your Text Encoder Can be Image Quality Controller}

\author{%
Yanyu Li$^{1, 2}$ \quad
Xian Liu$^{1}$ \quad
Anil Kag$^{1}$ \quad
Ju Hu$^1$  \quad
Yerlan Idelbayev$^1$ \\
\quad
Dhritiman Sagar$^1$ 
\quad
Yanzhi Wang$^2$ 
\quad
Sergey Tulyakov$^1$ 
\quad
Jian Ren$^{1}$
\\
$^1$Snap Inc. \quad  $^2$Northeastern University 
}

\newcommand{\modelname}{TextCraftor}

\begin{document}
\twocolumn[{
\renewcommand\twocolumn[1][]{#1}

\maketitle
\begin{center}
    \centering
    \captionsetup{type=figure}
    \includegraphics[width=\textwidth,height=6cm]{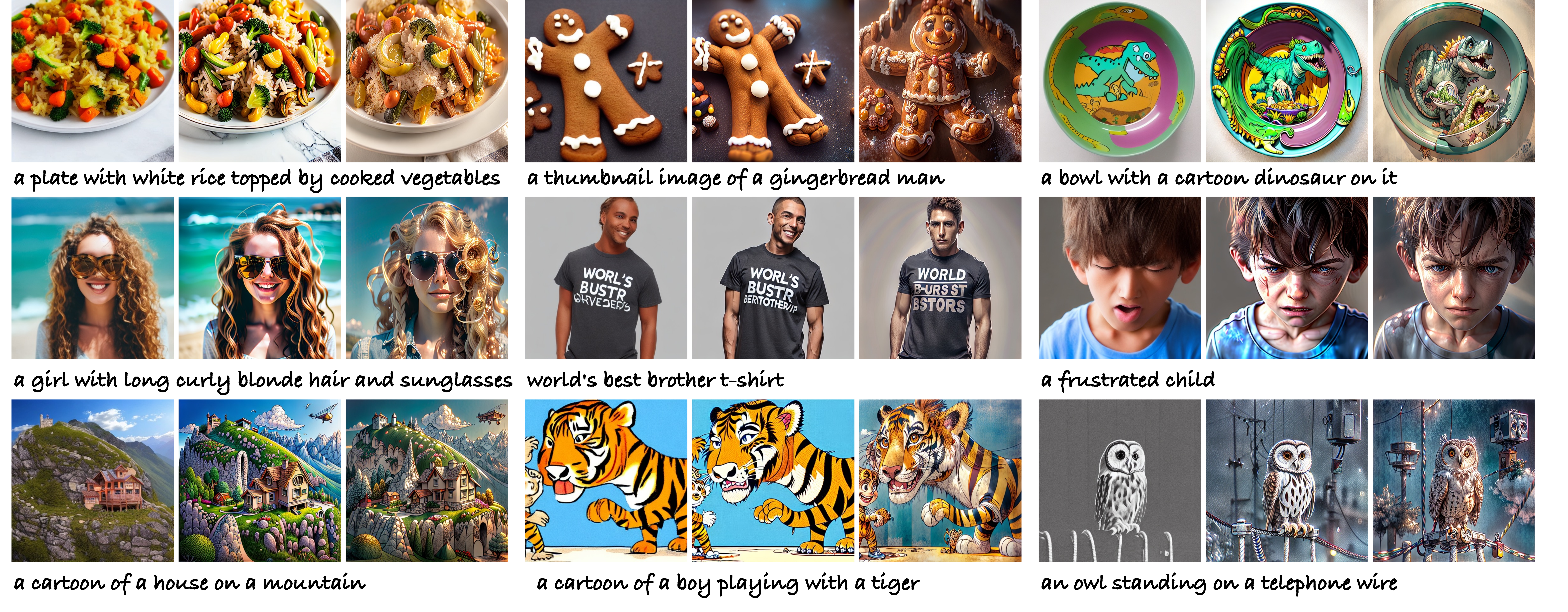}
    \vspace{-6mm}
    \caption{\textbf{Example generated images.} For each prompt, we show images generated from three different models, which are SDv1.5, \modelname, \modelname~+ UNet, listed from left to right. The random seed is fixed for all generation results.}{
    }
    \label{fig:teasor}
\end{center}
}]

\begin{abstract}
Diffusion-based text-to-image generative models, \eg, Stable Diffusion, have revolutionized the field of content generation, enabling significant advancements in areas like image editing and video synthesis.
Despite their formidable capabilities, these models are not without their limitations. 
It is still challenging to synthesize an image that aligns well with the input text, and multiple runs with carefully crafted prompts are required to achieve satisfactory results.
To mitigate these limitations, numerous studies have endeavored to fine-tune the pre-trained diffusion models, \ie., UNet, utilizing various technologies. 
Yet, amidst these efforts, a pivotal question of text-to-image diffusion model training has remained largely unexplored: 
\emph{\textbf{
Is it possible and feasible to fine-tune the text encoder to improve the performance of text-to-image diffusion models? 
}}
Our findings reveal that, instead of replacing the CLIP text encoder used in Stable Diffusion with other large language models, we can enhance it through our proposed fine-tuning approach, TextCraftor, leading to substantial improvements in quantitative benchmarks and human assessments. 
Interestingly, our technique also empowers controllable image generation through the interpolation of different text encoders fine-tuned with various rewards.
We also demonstrate that TextCraftor is orthogonal to UNet finetuning, and can be combined to further improve generative quality. 
\end{abstract}    
\section{Introduction}\label{sec:intro}
Recent breakthroughs in text-to-image diffusion models have brought about a revolution in content generation~\cite{ho2020denoising,song2020score,dhariwal2021diffusion,ldm,liu2023hyperhuman}. 
Among these models, the open-sourced Stable Diffusion (SD) has emerged as the \emph{de facto} choice for a wide range of applications, including image editing, super-resolution, and video synthesis~\cite{glide,li2023snapfusion,ho2022imagen,saharia2022palette,chang2023muse,lugmayr2022repaint,zhang2023sine,saharia2022image,singer2022make}.
Though trained on large-scale datasets, SD still holds two major challenges. \emph{First}, it often produces images that do not align well with the provided prompts~\cite{xue2023raphael,chen2023pixartalpha}.
\emph{Second}, generating visually pleasing images frequently requires multiple runs with different random seeds and manual prompt engineering~\cite{witteveen2022investigating,gu2023systematic}.
To address the \emph{first} challenge, prior studies explore the substitution of the CLIP text encoder~\cite{radford2021learning} used in SD with other large language models like T5~\cite{imagen,chung2022scaling}. Nevertheless, the large T5 model has an order of magnitude more parameters than CLIP, resulting in additional storage and computation overhead. 
In tackling the \emph{second} challenge, existing works fine-tune the pre-trained UNet from SD on paired image-caption datasets with reward functions~\cite{imagereward,clark2023directly,prabhudesai2023aligning}. Nonetheless, models trained on constrained datasets may still struggle to generate high-quality images for unseen prompts.

Stepping back and considering the pipeline of text-to-image generation, the text encoder and UNet should \emph{both} significantly influence the quality of the synthesized images. Despite substantial progress in enhancing the UNet model~\cite{si2023freeu,he2023scalecrafter}, limited attention has been paid to improving the text encoder.  This work aims to answer a pivotal question: \emph{Can fine-tuning a pre-trained text encoder used in the generative model enhance performance, resulting in better image quality and improved text-image alignment?}

To address this challenge, we propose \emph{\modelname}, an end-to-end fine-tuning technique to enhance the pre-trained text encoder. 
Instead of relying on paired text-image datasets, we demonstrate that reward functions (\emph{e.g.}, models trained to automatically assess the image quality like aesthetics model~\cite{schuhmann2023aestheticpredictor}, or text-image alignment assessment models~\cite{pickscore, hpsv2}) can be used to improve text-encoder in a differentiable manner. 
By only necessitating text prompts during training, \emph{\modelname} enables the on-the-fly synthesis of training images and alleviates the burden of storing and loading large-scale image datasets. 
We summarize our findings and contributions as follows:
\begin{itemize}
    \item We demonstrate that for a well-trained text-to-image diffusion model, fine-tuning text encoder is a buried gem, and can lead to significant improvements in image quality and text-image alignment (as in Fig.~\ref{fig:teasor} \& \ref{fig:main_results}). Compared with using larger text encoders, \emph{e.g.}, SDXL, \emph{TextCraftor} does not introduce extra computation and storage overhead. Compared with prompt engineering, \emph{TextCraftor} reduces the risks of generating irrelevant content. 
    \item {We introduce an effective and stable text encoder fine-tuning pipeline supervised by public reward functions. The proposed alignment constraint preserves the capability and generality of the large-scale CLIP-pretrained text encoder, making \emph{TextCraftor} the first generic reward fine-tuning paradigm among concurrent arts. Comprehensive evaluations on public benchmarks and human assessments demonstrate the superiority of \emph{TextCraftor}. }
    \item {We show that the textual embedding from different fine-tuned and original text encoders can be interpolated to achieve more diverse and controllable style generation. Additionally, \emph{TextCraftor} is orthogonal to UNet finetuning. We further show quality improvements by subsequently fine-tuning UNet with the improved text encoder.}
\end{itemize}

\section{Related Works}\label{sec:related}

\noindent\textbf{Text-to-Image Diffusion Models.} Recent efforts in the synthesis of high-quality, high-resolution images from natural language inputs have showcased substantial progress~\cite{ldm,ediffi}. 
Diverse investigations have been conducted to improve model performance by employing various network architectures and training pipelines, such as GAN-based approaches~\cite{kang2023scaling}, auto-regressive models~\cite{menick2018generating,yu2022scaling}, and diffusion models~\cite{karras2022elucidating,sohl2015deep,song2019generative,ho2020denoising,song2020score}.
Since the introduction of the Stable Diffusion models and their state-of-the-art performance in image generation and editing tasks, they have emerged as the predominant choice~\cite{ldm}. Nevertheless, they exhibit certain limitations.
For instance, the generated images may not align well with the provided text prompts~\cite{xue2023raphael}. Furthermore, achieving high-quality images may necessitate extensive prompt engineering and multiple runs with different random seeds~\cite{witteveen2022investigating,gu2023systematic}.
To address these challenges, one potential improvement involves replacing the pre-trained CLIP text-encoder~\cite{radford2021learning} in the Stable Diffusion model with T5~\cite{chung2022scaling} and fine-tuning the model using high-quality paired data~\cite{imagen,dai2023emu}. However, it is crucial to note that such an approach incurs a substantial training cost. Training the Stable Diffusion model alone from scratch demands considerable resources, equivalent to $6,250$ A100 GPUs days~\cite{chen2023pixartalpha}. This work improves pre-trained text-to-image models while significantly reducing computation costs.

\noindent\textbf{Automated Performance Assessment of Text-to-Image Models.} Assessing the performance of text-to-image models has been a challenging problem. Early methods use automatic metrics like FID to gauge image quality and CLIP scores to assess text-image alignment~\cite{dalle,dalle2}. 
However, subsequent studies have indicated that these scores exhibit limited correlation with human perception~\cite{podell2023sdxl}.
To address such discrepancies, recent research has delved into training models specifically designed for evaluating image quality for text-to-image models. Examples include ImageReward~\cite{imagereward}, PickScore~\cite{pickscore}, and human preference scores~\cite{wu2023human, hpsv2}, which leverage human-annotated images to train the quality estimation models. 
In our work, we leverage these models, along with an image aesthetics model~\cite{schuhmann2023aestheticpredictor}, as reward functions for enhancing visual quality and text-image alignment for the text-to-image diffusion models.

\noindent\textbf{Fine-tuning Diffusion Models with Rewards.} 
In response to the inherent limitations of pre-trained diffusion models, various strategies have been proposed to elevate generation quality, focusing on aspects like image color, composition, and background~\cite{lee2023aligning,dong2023raft}. One direction utilizes
reinforcement learning to fine-tune the diffusion model~\cite{black2023training,2023DPOK}. Another area fine-tunes the diffusion models with reward function in a differentiable manner~\cite{imagereward}. Following this trend, later studies extend the pipeline to trainable LoRA weights~\cite{hu2021lora} with the text-to-image models~\cite{clark2023directly,prabhudesai2023aligning}. In our work, we delve into the novel exploration of fine-tuning the text-encoder using reward functions in a differentiable manner, a dimension that has not been previously explored.

\noindent\textbf{Improving Textual Representation.} Another avenue of research focuses on enhancing user-provided text to generate images of enhanced quality. Researchers use large language models, such as LLAMA~\cite{touvron2023llama}, to refine or optimize text prompts~\cite{hao2022optimizing,pryzant2023automatic,zhong2023adapter}.
By improving the quality of prompts, the text-to-image model can synthesize higher-quality images. However, the utilization of additional language models introduces increased computational and storage demands. This study demonstrates that by fine-tuning the text encoder, the model can gain a more nuanced understanding of the given text prompts, obviating the need for additional language models and their associated overhead.

\section{Method}\label{sec:method}

\begin{figure*}[]
    \centering
    \includegraphics[width=0.82\linewidth]{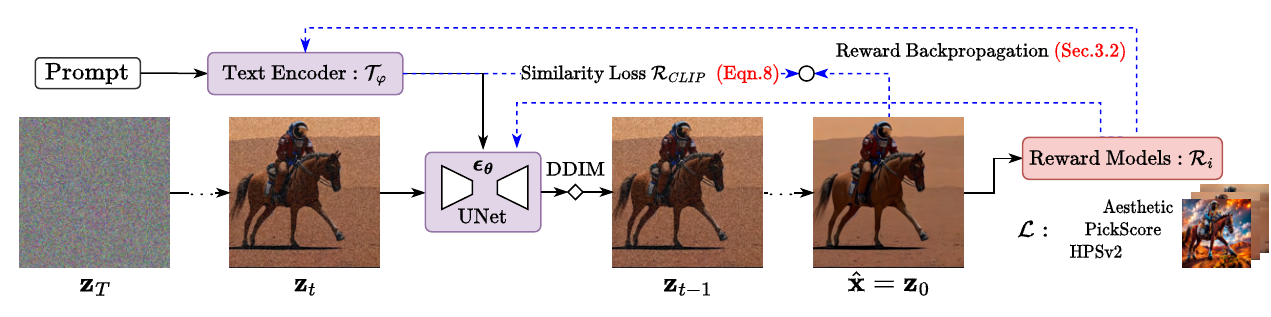}  
    \caption{\textbf{Overview of \modelname}, an end-to-end text encoder fine-tuning paradigm based on prompt data and reward functions. The text embedding is forwarded into the DDIM denoising chain to obtain the output image and compute the reward loss, 
    then we backward to update the parameters of the text encoder (and optionally UNet) by maximizing the reward.}
    \label{fig:pipeline}
\end{figure*}

\subsection{Preliminaries of Latent Diffusion Models} \label{preliminaries}

\noindent\textbf{Latent Diffusion Models.}
Diffusion models convert the real data distribution \eg, images, into a noisy distribution, \eg, Gaussian distribution, and can reverse such a process to for randomly sampling~\cite{sohl2015deep}. To reduce the computation cost, \eg, the number of denoising steps, latent diffusion model (LDM) proposes to conduct the denoising process in the latent space~\cite{ldm} using a UNet~\cite{ronneberger2015unet,ho2020denoising}, where real data is encoded through variational autoencoder (VAE)~\cite{kingma2013auto,rezende2014stochastic}. 
The latent is then decoded into an image during inference time. 
LDM demonstrates promising results for text-conditioned image generation. 
Trained with large-scale text-image paired datasets~\cite{schuhmann2022laion}, a series of LDM models, namely, Stable Diffusion~\cite{ldm}, are obtained. The text prompts are processed by a pre-trained text encoder, which is the one from CLIP~\cite{radford2021learning} used by Stable Diffusion, to obtain textual embedding as the condition for image generation.
In this work, we use the Stable Diffusion as the baseline model to conduct most of our experiments, as it is widely adopted in the community for various tasks. 

Formally, let ($\mathbf{x}$, $\mathbf{p}$) be the real-image and prompt data pair (for notation simplicity, $\mathbf{x}$ also represents the data encoded by VAE) drawn from the distribution $p_{\text{data}}(\mathbf{x}, \mathbf{p})$, $\hat{\bm{\epsilon}}_{\bm{\theta}}(\cdot)$ be the diffusion model with parameters $\bm{\theta}$, ${\mathcal{T}}_{\bm{\varphi}}(\cdot)$ be the text encoder parameterized by $\bm{\varphi}$, training the text-to-image LDM under the objective of noise prediction can be formulated as follows~\cite{sohl2015deep,ho2020denoising,song2020score}:
\begin{equation}\label{eqn:train_loss}
\small
    \min_{\bm{\theta}} \; \mathbb{E}_{t\sim U[0, 1],(\mathbf{x},\mathbf{p})\sim p_{\text{data}}(\mathbf{x},\mathbf{p}),\bm{\epsilon}\sim\mathcal{N}(\mathbf{0}, \mathbf{I})} \; \lvert\lvert\hat{ \bm{\epsilon}}_{\bm{\theta}}(t, \mathbf{z}_t, \mathbf{c}) -\bm{\epsilon} \rvert\rvert_2^2,
\end{equation}
where $\bm{\epsilon}$ is the ground-truth noise; $t$ is the time step; $\mathbf{z}_t=\alpha_t\mathbf{x}+\sigma_t\bm{\epsilon}$ is the noised sample with $\alpha_t$ represents the signal and $\sigma_t$ represents the noise, that both decided by the scheduler; and $\mathbf{c}$ is the textual embedding such that $\mathbf{c} = \mathcal{T}_{\bm{\varphi}} (\mathbf{p})$.

During the training of SD models, the weights of text encoder $\mathcal{T}$ are fixed. However, the text encoder from CLIP model is optimized through the contrastive objective between text and images. Therefore, it does not necessarily learn the semantic meaning of the prompt, resulting the generated image might not align well with the given prompt using such a text encoder. In Sec.~\ref{sec:text-encoder-ft}, we introduce the technique of improving the text encoder without using the text and image contrastive pre-training in CLIP~\cite{radford2021learning}.

\noindent\textbf{Denoising Scheduler -- DDIM.}
After a text-to-image diffusion model is trained, we can sample Gaussian noises for the same text prompt using numerous samplers, such as DDIM~\cite{ddim}, that iteratively samples from $t$ to its previous step $t'$ with the following denoising process, until $t$ becomes $0$:
\begin{equation} \label{eq:ddim}
\small
\mathbf{z}_{t^\prime}=\alpha_{t^\prime} \frac{\mathbf{z}_t - \sigma_t \hat{\bm{\epsilon}}_{\bm{\theta}}(t, \mathbf{z}_t, \mathbf{c})}{\alpha_t} + \sigma_{t^\prime} \hat{\bm{\epsilon}}_{\bm{\theta}}(t, \mathbf{z}_t, \mathbf{c}).
\end{equation}


\noindent\textbf{Classifier-Free Guidance.} One effective approach to improving the generation quality during the sampling stage is the classifier-free guidance (CFG)~\cite{cfg}. By adjusting the guidance scale $w$ in CFG, we can further balance the trade-off between the fidelity and the text-image alignment of the synthesized image. Specifically, for the process of text-conditioned image generation, by letting  $\varnothing$ denote the null text input, classifier-free guidance can be defined as follows:
\begin{equation}\label{eqn:cfg}
\small
    \hat{\bm{\epsilon}}
    = w\hat{\bm{\epsilon}}_{\bm{\theta}}(t, \mathbf{z}_t, \mathbf{c}) - (w-1) \hat{\bm{\epsilon}}_{\bm{\theta}}(t, \mathbf{z}_t, \varnothing).
\end{equation}


\subsection{Text Encoder Fine-tuning with Reward Propagation}\label{sec:text-encoder-ft}


We introduce and experiment with two techniques for fine-tuning the text encoder by reward guidance. 

\subsubsection{Directly Fine-tuning with Reward}\label{sec:direct_finetune}
Recall that for a normal training process of diffusion models, we sample from real data and random noise to perform forward diffusion: $\mathbf{z}_t=\alpha_t\mathbf{x}+\sigma_t\bm{\epsilon}$, upon which the denoising UNet, $\hat{\bm{\epsilon}}_{\bm{\theta}}(\cdot)$,
makes its (noise) prediction. 
Therefore, instead of calculating $\mathbf{z}_{t'}$ as in Eqn.~\ref{eq:ddim},
we can alternatively predict the original data as follows~\cite{ddim}, 
\begin{equation} \label{eq:xpred}
\small
\mathbf{\hat x} = \frac{\mathbf{z}_t - \sigma_t \hat{\bm{\epsilon}}_{\bm{\theta}}(t, \mathbf{z}_t, \mathcal{T}_{\bm{\varphi}}(\mathbf{p}))}{\alpha_t},
\end{equation}
where $\mathbf{\hat x}$ is the estimated real sample, which is an image for the text-to-image diffusion model. 
Our formulation works for both pixel-space and latent-space diffusion models, where in latent diffusion, $\hat{\mathbf{x}}$ is actually post-processed by the VAE decoder before feeding into reward models. 
Since the decoding process is also differentiable, for simplicity, we omit this process in formulations and simply refer $\hat{\mathbf{x}}$ as the predicted image. 
With $\mathbf{\hat x}$ in hand, we are able to utilize public reward models, denoted as $\mathcal{R}$, to assess the quality of the generated image.
Therefore, to improve the text encoder used in the diffusion model, we can optimize its weights, \ie, $\bm{\varphi}$ in $\mathcal{T}$,  with the learning objective as maximizing the quality scores predicted by reward models.

More specifically, we employ both image-based reward model $\mathcal{R}(\mathbf{\hat x})$, \ie, Aesthetic score predictor~\cite{schuhmann2023aestheticpredictor}, and text-image alignment-based reward models $\mathcal{R}(\mathbf{\hat x}, \mathbf{p})$, \ie, HPSV2~\cite{hpsv2} and PickScore~\cite{pickscore}. 
Consequently, the loss function for maximizing the reward scores can be defined as follows,
\begin{equation} \label{eq:loss_direct}
\small
\begin{aligned}
        \mathcal{L}(\varphi) &= - {\mathcal{R}} (\hat{\mathbf{x}}, \cdot / \mathbf{p}) \\
        &= - {\mathcal{R}} (\frac{\mathbf{z}_t - \sigma_t {\bm{\epsilon}}_{\bm{\theta}}(t, \mathbf{z}_t, {\mathcal{T}}_\varphi(\mathbf{p}))}{\alpha_t}, \cdot / \mathbf{p}).
\end{aligned}
\end{equation}
Note that when optimizing Eqn.~\ref{eq:loss_direct}, the weights for all reward models and the UNet model are fixed, while only the weights in the CLIP text encoder are modified. 

\noindent\textbf{Discussion.}
Clearly, directly fine-tuning shares a similar training regime with regular training of diffusion models, where we are ready to employ text-image paired data ($\mathbf{x}, \mathbf{p}$) and predict reward by converting predicted noise into the predicted real data $\hat{\mathbf{x}}$. 
However, considering the very beginning (noisy) timesteps, the estimated $\hat{\mathbf{x}}$ can be \emph{inaccurate} and \emph{less reliable}, making the predicted reward less meaningful.
Instead of utilizing $\hat{\mathbf{x}}$, Liu \etal \cite{liu2023more} propose to fine-tune the reward models to enable a noisy latent ($\mathbf{z}_t$) aware score prediction, which is out of the scope of this work. 
For the best flexibility and sustainability of our method, we only investigate publicly available reward models, thus we directly employ $\hat{\mathbf{x}}$ prediction. 
We discuss the performance of direct finetuning in Section.~\ref{sec:exp}.

\subsubsection{Prompt-Based Fine-tuning}

\begin{algorithm}[]
\small
\caption{Prompt-Based Reward Finetuning}
\label{algorithm:prompt-finetune}
\begin{algorithmic}
\Require  Pretrained UNet
$\hat{\bm{\epsilon}}_{\bm\theta}$; pretrained text encoder $\mathcal{T}_{\varphi}$;
prompt set: $\mathbb P \{\mathbf{p}\}$.
\Ensure $\mathcal{T}_{\varphi}$ (optionally
$\hat{\bm{\epsilon}}_{\bm\theta}$ if fine-tuning UNet) converges and maximizes $\mathcal{L}_{total}$. 

\State $\rightarrow$ \textbf{Perform text encoder fine-tuning}.
\State Freeze UNet $\hat{\bm{\epsilon}}_{\bm\theta}$ and reward models $\mathcal{R}_i$, activate $\mathcal{T}_{\varphi}$. 
\While{$\mathcal{L}_{total}$ not converged}
\State Sample $\mathbf{p}$ from $\mathbb{P}$; $t=T$
\While{$t>0$}
\State $\mathbf{z}_{t-1} \leftarrow \alpha_{t'} \frac{\mathbf{z}_t - \sigma_t \hat{\bm{\epsilon}}_{\bm{\theta}}(t, \mathbf{z}_t, \mathcal{T}_{\varphi}(\mathbf{p}))}{\alpha_t} + \sigma_{t'} \hat{\bm{\epsilon}}_{\bm{\theta}}(t, \mathbf{z}_t, \mathcal{T}_{\varphi}(\mathbf{p}))$
\EndWhile
\State $\hat{\mathbf{x}} \leftarrow \mathbf{z}_0$
\State $    \mathcal{L}_{total} \leftarrow  - \sum_i \gamma_i {\mathcal{R}_i} (\hat{\mathbf{x}}, \cdot / \mathbf{p}).$
\State Backward $\mathcal{L}_{total}$ and update $\mathcal{T}_{\varphi}$ for last K steps. 

\EndWhile

\State $\rightarrow$ \textbf{Perform UNet finetuning}.
\State Freeze $\mathcal{T}_{\varphi}$ and reward models $\mathcal{R}_i$, activate UNet $\hat{\bm{\epsilon}}_{\bm\theta}$. 
\State Repeat the above reward training until converge. 


\end{algorithmic}   

\end{algorithm}

As an alternative way to overcome the problem of the inaccurate $\hat{\mathbf{x}}$ prediction, given a specific text prompt $\mathbf{p}$ and an initial noise $\mathbf{z}_T$, we can iteratively solve the denoising process in Eqn.~\ref{eq:ddim} to get $\hat{\mathbf{x}} = \mathbf{z}_0$, 
which can then be substituted to Eqn.~\ref{eq:loss_direct} to compute the reward scores. 
Consequently, we are able to precisely predict $\hat{\mathbf{x}}$, and also eliminate the need for paired text-image data and perform the reward fine-tuning with \emph{only prompts} and a pre-defined denoising schedule, \ie, 25-steps DDIM in our experiments. 
Since each timestep in the training process
is differentiable, the gradient to update $\bm{\varphi}$ in $\mathcal{T}$ can be calculated through chain rule as follows, 
\begin{equation} \label{eq:prompt-finetune}
\tiny
    \frac{\partial \mathcal{L}}{\partial \varphi} = - \frac{\partial {\mathcal{R}}}{\partial \hat{\mathbf{x}}} \cdot  \prod_{t=0}^{t} \frac{\partial [\alpha_{t^\prime} \frac{\mathbf{z}_t - \sigma_t \hat{\bm{\epsilon}}_{\bm{\theta}}(t, \mathbf{z}_t, {\mathcal{T}}_{\varphi}(\mathbf{p}))}{\alpha_t} + \sigma_{t^\prime} \hat{\bm{\epsilon}}_{\bm{\theta}}(t, \mathbf{z}_t, {\mathcal{T}}_{\varphi}(\mathbf{p}))]}{\partial {\mathcal{T}}_{\varphi}(\mathbf{p})} \cdot \frac{\partial {\mathcal{T}}_{\varphi}(\mathbf{p})}{\partial \varphi}.
\end{equation}
It is notable that solving Eqn.~\ref{eq:prompt-finetune} is memory infeasible for early (noisy) timesteps, \ie, $t=\{T, T-1, ...\}$, as the computation graph accumulates in the backward chain. 
We apply gradient checkpointing~\cite{chen2016training} to trade memory with computation. 
Intuitively, the intermediate results are re-calculated on the fly, thus the training can be viewed as solving one step at a time. 
Though with gradient checkpointing, we can technically train the text encoder with respect to each timestep, early steps still suffer from gradient explosion and vanishing problems in the long-lasting accumulation~\cite{clark2023directly}. 
We provide a detailed analysis of step selection in Section.~\ref{ablation}. 
The proposed prompt-based reward finetuning is further illustrated in Fig.~\ref{fig:pipeline} and Alg.~\ref{algorithm:prompt-finetune}.

\subsection{Loss Function}
We investigate and report the results of using multiple reward functions, where the reward losses $\mathcal{L}_{total}$ can be weighted by $\gamma$ and linearly combined as follows, 
\begin{equation}\label{eq:loss_total}
\small
    \mathcal{L}_{total} = \sum_i \mathcal{L}_i = - \sum_i \gamma_i {\mathcal{R}_i} (\hat{\mathbf{x}}, \cdot / \mathbf{p}).
\end{equation}
Intuitively, we can arbitrarily combine different reward functions with various weights. 
However, as shown in Fig.~\ref{fig:ablate-clip-constraint}, some reward functions are by nature limited in terms of their capability and training scale. 
As a result, fine-tuning with only one reward can result in catastrophic forgetting and mode collapse. 

To address this issue, recent works~\cite{imagereward,black2023training} mostly rely on careful tuning, including focusing on a specific subdomain, \eg, human and animals~\cite{prabhudesai2023aligning}, and early stopping~\cite{clark2023directly}.
Unfortunately, this is not a valid approach in the generic and large-scale scope. 
In this work, we aim at enhancing generic models and eliminating human expertise and surveillance. 

To achieve this,
we set CLIP space similarity as an always-online constraint as follows, 
\begin{equation}\small
    \mathcal{R}_{CLIP} = \texttt{cosine-sim}(\mathcal{I}(\hat{\mathbf{x}}), \mathcal{T}_{\varphi}(\mathbf{p})),
    \label{eq:clip-constraint}
\end{equation}
and ensure $\gamma_\text{CLIP} > 0$ in Eqn.~\ref{eq:loss_total}. 
Specifically, we maximize the cosine similarity between the textual embeddings and image embeddings.
The textual embedding is obtained in forward propagation, while the image embedding is calculated by sending the predicted image $\hat{\mathbf{x}}$ to the image encoder $\mathcal{I}$ of CLIP. 
The original text encoder $\mathcal{T}_{\varphi}$ is pre-trained in large-scale contrastive learning paired with the image encoder $\mathcal{I}$ \cite{radford2021learning}. 
As a result, the CLIP constraint preserves the coherence of the fine-tuned text embedding and the original image domain, ensuring capability and generalization. 

\subsection{UNet Fine-tuning with Reward Propagation}\label{sec:unet-ft}

The proposed fine-tuning approach for text encoder is orthogonal to UNet reward fine-tuning~\cite{clark2023directly,prabhudesai2023aligning}, meaning that the text-encoder and UNet can be optimized under similar learning objectives to further improve performance. 
Note that our fine-tuned text encoder can seamlessly fit the pre-trained UNet in Stable Diffusion, and can be used for other downstream tasks besides text-to-image generation. 
To preserve this characteristic and avoid domain shifting, we fine-tune the UNet by freezing the finetuned text encoder ${\mathcal{T}}_{\varphi}$. The learning objective for UNet is similar as Eqn.~\ref{eq:prompt-finetune}, where we optimize parameters $\bm{\theta}$ of $\hat{\bm{\epsilon}}_{\bm{\theta}}$, instead of $\bm{\varphi}$.

\section{Experiments}\label{sec:exp}

\begin{figure*}[]
    \centering
    \includegraphics[width=0.95\linewidth]{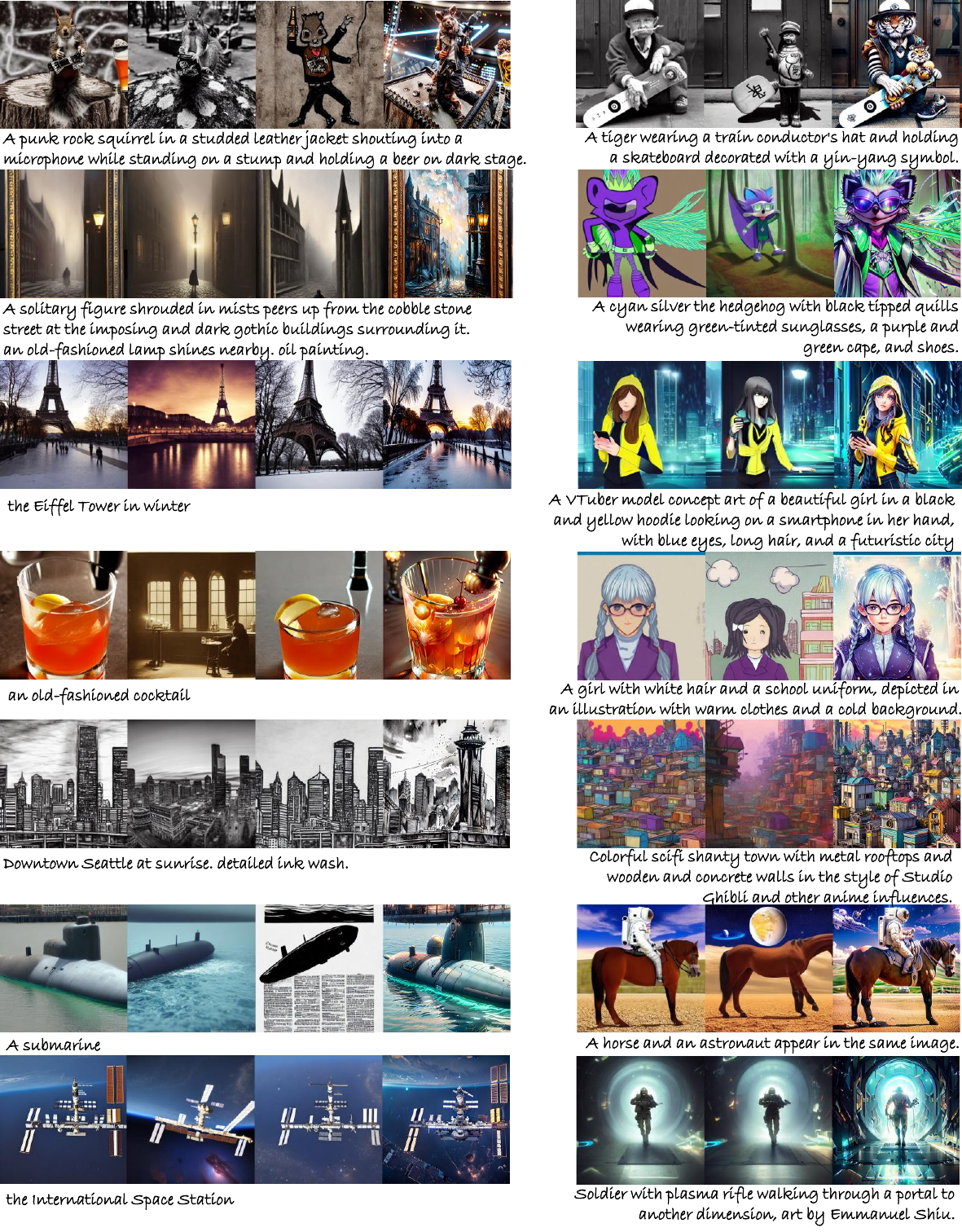}  
    \caption{\textbf{Qualitative visualizations.} \emph{\textbf{Left}}: generated images on Parti-Prompts, in the order of SDv1.5, prompt engineering, DDPO, and \modelname. \emph{\textbf{Right}}: examples from HPSv2, ordered as SDv1.5, prompt engineering, and \modelname. }
    \label{fig:main_results}
\end{figure*}

\noindent\textbf{Reward Functions.} We use image-based aesthetic predictor~\cite{schuhmann2023aestheticpredictor}, text-image alignment-based CLIP predictors, (\ie, Human Preference Score v2 (HPSv2)~\cite{hpsv2} and PickScore~\cite{pickscore}), and CLIP model~\cite{radford2021learning}. 
We adopt the improved (v2) version of the aesthetic predictor that is trained on $176,000$ image-rating pairs. 
The predictor estimates a quality score ranging from $1$ to $10$, where larger scores indicate higher quality. 
HPSv2 is a preference prediction model trained on a large-scale well-annotated dataset of human choices, with $798$K preference annotations and $420$K images. 
Similarly, PickScore~\cite{pickscore} is a popular human preference predictor trained with over half a million samples. 

\noindent\textbf{Training Datasets.} We perform training on OpenPrompt\footnote{\url{https://github.com/krea-ai/open-prompts}} dataset, which includes more than $10$M high quality prompts for text-to-image generation. 
For direct finetuning, we use the public LAION-2B dataset with conventional pre-processing, \emph{i.e.}, filter out NSFW data, resize and crop images to $512^2$px, and use Aesthetics$>5.0$ images. 

\noindent\textbf{Experimental Settings.} We conduct experiments with the latest PyTorch~\cite{NEURIPS2019_9015} and HuggingFace Diffusers\footnote{\url{https://github.com/huggingface/diffusers}}. 
We choose Stable Diffusion v1.5 (SDv1.5)~\cite{ldm} as the baseline model, as it performs well in real-world human assessments with appealing model size and computation than other large diffusion models~\cite{podell2023sdxl}. We fine-tune the ViT-L text encoder of SDv1.5, which takes $77$ tokens as input and outputs an embedding with dimension $768$. 
The fine-tuning is done on $8$ NVIDIA A100 nodes with 8 GPUs per node, using AdamW optimizer~\cite{loshchilov2017decoupled} and a learning rate of $10^{-6}$. 
We set CFG scale to 7.5 in all the experiments. 

\noindent\textbf{Comparison Approaches.} We compare our method with the following approaches.
\begin{itemize}
    \item \emph{Pre-trained} text-to-image models that include SDv1.5, SDv2.0, SDXL Base 0.9, and DeepFloyd-XL.
    \item \emph{Direct fine-tuning} that is described in Sec.~\ref{sec:direct_finetune}.
    \item \emph{Reinforcement learning-based approach} that optimize the diffusion model using reward functions~\cite{black2023training}.
    \item \emph{Prompt engineering}. From the scope of the enhancement of text information, prompt engineering~\cite{witteveen2022investigating,gu2023systematic} can be considered as a counterpart of our approach. By extending and enriching the input prompt with more detailed instructions, \eg, using words like 4K, photorealistic, ultra sharp, etc., the output image quality could be greatly boosted.
However, prompt engineering requires case-by-case human tuning, which is not appealing in real-world applications. 
Automatic engineering method\footnote{\url{https://huggingface.co/daspartho/prompt-extend}} employs text generation models to enhance the prompt, while the semantic coherence might not be guaranteed. 
We experiment and compare with automatic prompt engineering on both quantitative and qualitative evaluations.

\end{itemize}

\noindent\textbf{Quantitative Results.}
We report the results with different training settings (the CLIP constraint is utilized under all the settings of our approach) on two datasets:
\begin{itemize}
    \item We report zero-shot evaluation results for the score of three rewards on Parti-Prompts~\cite{yu2022scaling}, which contains $1632$ prompts with various categories, in Tab.~\ref{tab:parti}. We show experiments using a single reward function, \eg, Aesthetics, and the combination of all reward functions, \ie, denoted as All. We also fine-tune the UNet by freezing the fine-tuned text-encoder (\modelname~+ UNet in Tab.~\ref{tab:parti}). We evaluate different methods by forwarding the generated images (or the given prompt) to reward functions to obtain scores, where higher scores indicate better performance.
    \item  We report zero-shot results on the HPSv2 benchmark set, which contains $4$ subdomains of animation, concept art, painting, and photo, with $800$ prompts per category. 
    In addition to the zero-shot model trained with combined rewards (denote as All in Tab.~\ref{tab:hpsv2}),
    we train the model solely with HPSv2 reward to report the best possible scores TextCraftor can achieve on the HPSv2 benchmark. 
\end{itemize}


\noindent From the results, we can draw the following observations:
\begin{itemize}
    \item Compared to the pre-trained text-to-image models, \ie, SDv1.5 and SDv2.0, our \modelname~achieves significantly higher scores, \ie, Aesthetics, PickScore, and HPSv2, compared to the baseline SDv1.5. More interestingly, \modelname~ outperforms SDXL Base 0.9. and DeepFloyd-XL, which have much larger UNet and text encoder.
    \item Direct fine-tuning (described in Sec.~\ref{sec:direct_finetune}) can not provide reliable performance improvements. 
     \item Compared to prompt engineering, \modelname~obtains better performance, without necessitating human effort and ambiguity. We notice that the incurred additional information in the text prompt leads to lower alignment scores. 
    \item Compared to previous state-of-the-art DDPO~\cite{black2023training} that performs reward fine-tuning on UNet, we show that \emph{\modelname~+ UNet} obtains better metrics by a large margin. It is notable that DDPO is fine-tuned on subdomains, \eg, animals and humans, with early stopping, limiting its capability to generalize for unseen prompts. The proposed \modelname~is currently the first large-scale and generic reward-finetuned model. 
    \item Lastly, fine-tuning the UNet can further improve the performance, proving that \modelname~is orthogonal to UNet fine-tuning and can be combined to achieve significantly better performance.

\end{itemize}


\begin{table}[]
\small
\centering
\caption{\textbf{Comparison results on Parti-Prompts~\cite{yu2022scaling}.} We perform \modelname~fine-tuning on individual reward functions, including Aesthetics, PisckScore, and HPSv2, and the combination of all rewards to form a more generic model. }
\resizebox{1\linewidth}{!}{
\begin{tabular}{lcccc}
\toprule
Parti-1632  & Reward    & Aesthetics & PickScore & HPSv2  \\
\hline
SDXL Base 0.9  & -         &  5.7144   & 20.466   & 0.2783  \\
SDv2.0     & -         &  5.1675   &  18.893    & 0.2723  \\
SDv1.5     & -         & 5.2634    & 18.834    & 0.2703 \\
\hline
DDPO~\cite{black2023training}       & Aesthetic &    5.1424       &     18.790      &    0.2641    \\
DDPO~\cite{black2023training}       & Alignment &    5.2620       &     18.707      &    0.2676    \\
Prompt Engineering        & - &    5.7062       &     17.311      &    0.2599    \\
Direct Fine-tune (Sec.~\ref{sec:direct_finetune})        & All &    5.2880       &     18.750      &    0.2701    \\
\hline
\modelname        & Aesthetics & 5.5212   &  18.956   &  0.2670 \\
\modelname        & PickScore & 5.2662   &  19.023   &  0.2641 \\
\modelname        & HPSv2     & 5.4506    & 18.922    & 0.2800 \\
\modelname~(Text)      & All       & 5.8800    & 19.157    & 0.2805 \\
\modelname~(UNet)       & All       & 6.0062    & 19.281    & 0.2867 \\
\modelname~(Text+UNet) & All       & 6.4166    & 19.479    & 0.2900 \\
\bottomrule
\end{tabular}
}
\label{tab:parti}
\end{table}

\begin{table}[]
\small
\centering
\caption{\textbf{Comparison results on HPSv2 benchmark~\cite{hpsv2}.} In addition to the generic model, we report \modelname~fine-tuned solely on HPSv2 reward, denoted as \modelname~(HPSv2).
}
\resizebox{1\linewidth}{!}{
\begin{tabular}{lccccc}
\toprule
HPS-v2        & Animation & Concept Art & Painting & Photo  & \textbf{Average} \\
\hline
DeepFloyd-XL  & 0.2764    & 0.2683      & 0.2686   & 0.2775 & 0.2727  \\
SDXL Base 0.9 & 0.2842    & 0.2763      & 0.2760   & 0.2729 & 0.2773  \\
SDv2.0       & 0.2748    & 0.2689      & 0.2686   & 0.2746 & 0.2717  \\
SDv1.5       & 0.2721    & 0.2653      & 0.2653   & 0.2723 & 0.2688  \\
\hline
\modelname~(HPSv2)          & 0.2938    & 0.2919      & 0.2930   & 0.2851 & 0.2910  \\
\modelname~+ UNet (HPSv2)     & 0.3026    & 0.2995      & 0.3005   & 0.2907 &  0.2983 \\
\modelname~(All)       & 0.2829    & 0.2800      & 0.2797   & 0.2801 & 0.2807  \\
\modelname~+ UNet (All)    & 0.2885    & 0.2845      & 0.2851   & 0.2807 & 0.2847  \\
\bottomrule
\end{tabular}
}
\label{tab:hpsv2}
\end{table}

\begin{table}[]
\centering
\caption{\textbf{Human evaluation} on $1632$ Parti-Prompts~\cite{yu2022scaling}. 
Human annotators are given two images generated from different approaches and asked to choose the one that has better image quality and text-image alignment. Our approach obtains better human preference over all compared methods. 
}
\resizebox{\linewidth}{!}{
\begin{tabular}{lcccccc}
\toprule
Comparison Methods        & SDv1.5  & SDv2.0 & SDXL Base 0.9 & Prompt Eng. & DDPO Align. & DDPO Aes.  \\
\hline
Our Win Rate & 71.7\% & 81.7\% & 59.7\%& 81.3\% &56.7\%&66.2\% \\
\bottomrule
\end{tabular}}
\label{tab:human_eval}
\end{table}



\noindent\textbf{Qualitative Results.}
We demonstrate the generative quality of TextCraftor in Fig.~\ref{fig:teasor} and \ref{fig:main_results}.
Images are generated with the same noise seed for direct and fair comparisons. 
We show that with TextCraftor, the generation quality is greatly boosted compared to SDv1.5. 
Additionally, compared to prompt engineering, TextCraftor exhibits more reliable text-image alignment and rarely generates additional or irrelevant objects. 
Compared to DDPO~\cite{black2023training}, the proposed TextCraftor resolves the problem of mode collapse and catastrophic forgetting by employing text-image similarity as the constraint reward. 
We also show that fine-tuning the UNet models upon the TextCraftor enhanced text encoder can further boost the generation quality in the figure in the Appendix. 
From the visualizations, we observe that the reward fine-tuned models tend to generate more artistic, sharp, and colorful styles, which results from the preference of the reward models. 
When stronger and better reward predictors emerge in the future, TextCraftor can be seamlessly extended to obtain even better performance. 
Lastly, we provide a comprehensive human evaluation in Tab.~\ref{tab:human_eval}, proving the users prefer the images synthesized by \modelname.


\subsection{Controllable Generation}
Instead of adjusting reward weights $\gamma_i$ in Eqn.~\ref{eq:loss_total}, we can alternatively train dedicated text encoders optimized for each reward, and mix-and-match them in the inference phase for flexible and controllable generation. 

\begin{figure}[]
    \centering
    \includegraphics[width=\linewidth]{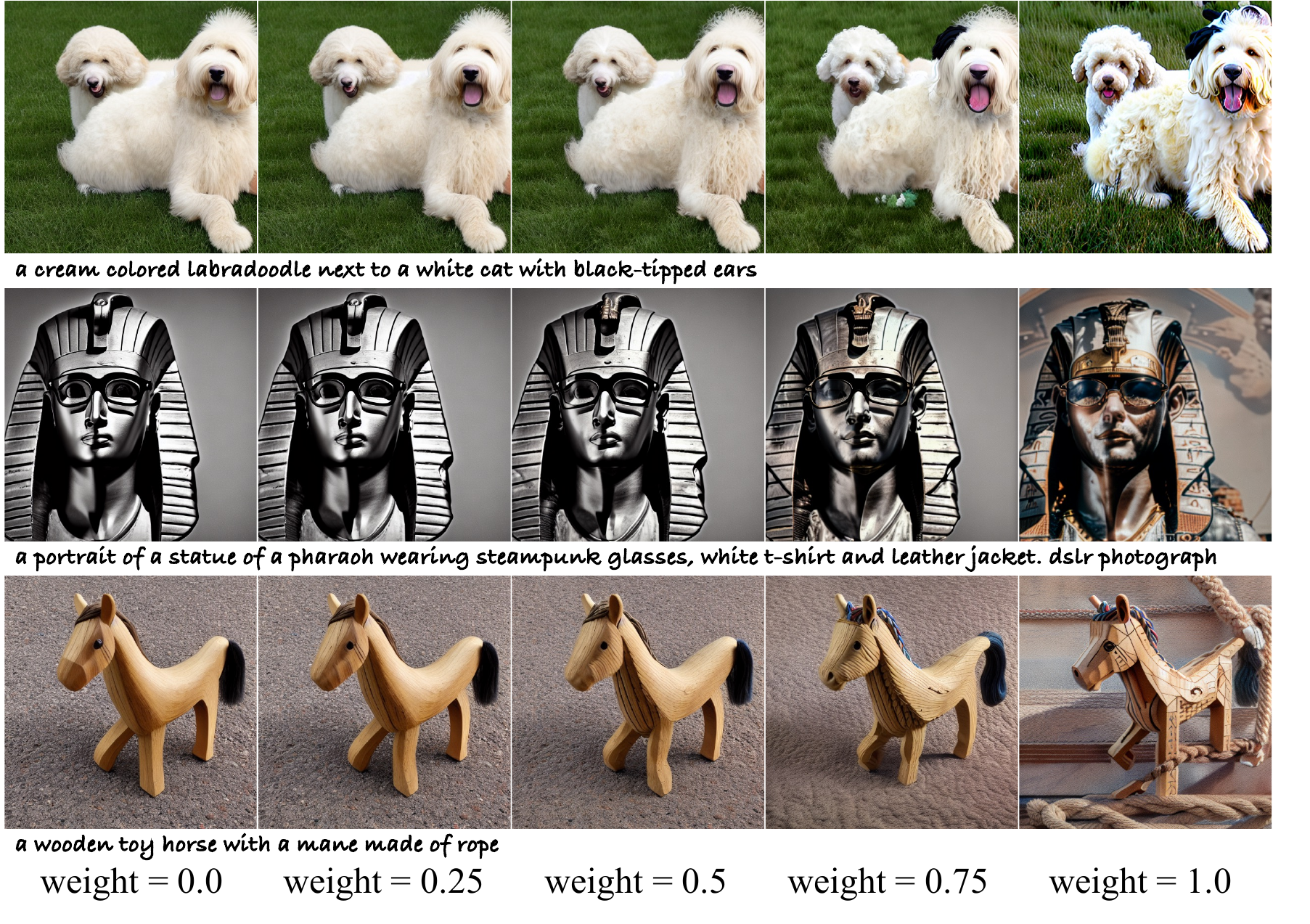}
    \caption{\textbf{Interpolation} between original text embedding (weight $0.0$) and the one from \modelname~(weight $1.0$)
    , demonstrating controllable generation.
    \emph{From top to bottom row}: \modelname~using HPSv2, PickScore, and Aesthetics as reward models. }
    \label{fig:weighting}
\end{figure}

\begin{figure}[]
    \centering
    \includegraphics[width=\linewidth]{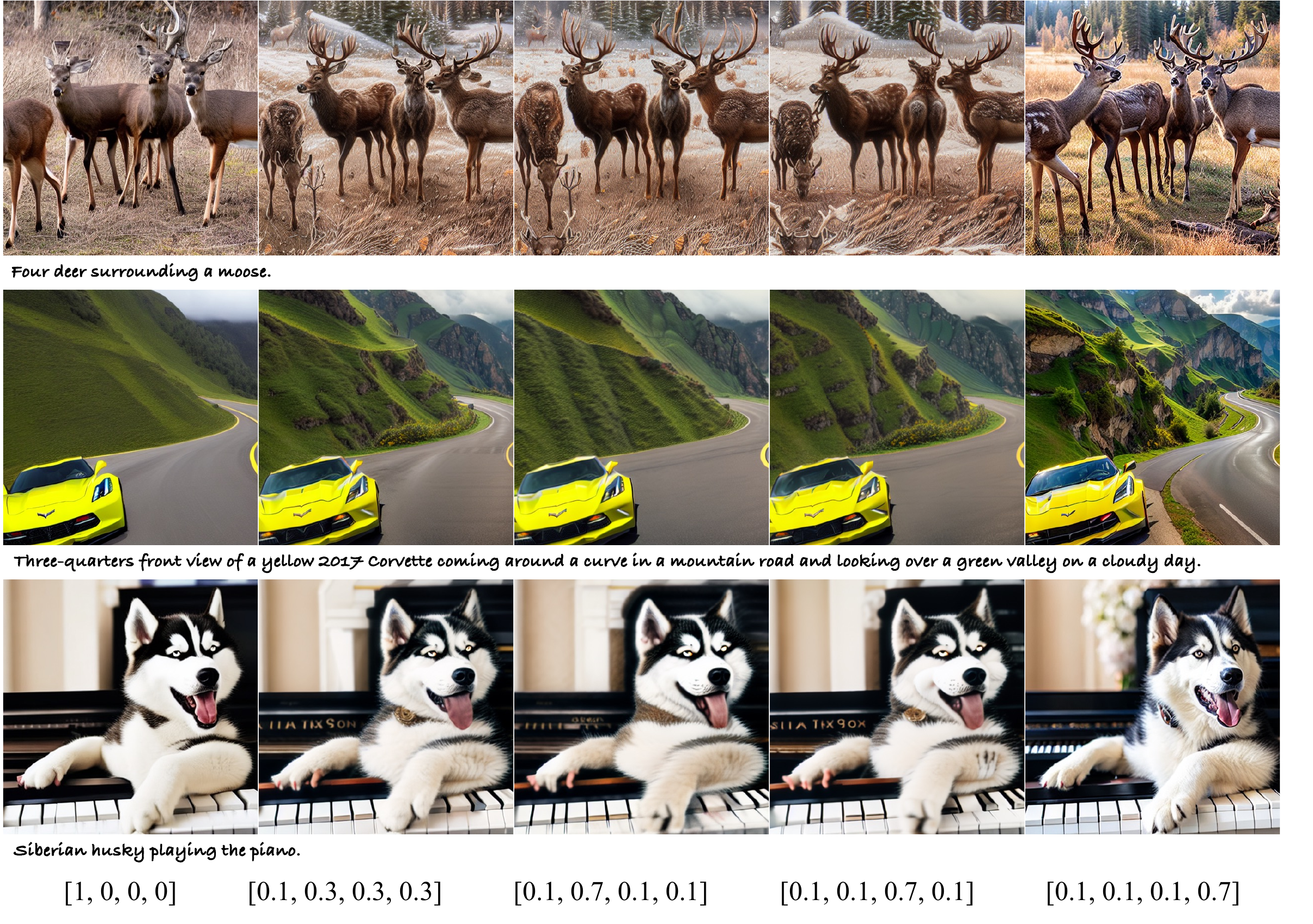}
    \caption{\textbf{Style mixing.} Text encoders fine-tuned from different reward models can collaborate and serve as style mixing. The weights listed at the bottom are used for combining text embedding from \{origin, Aesthetics, PickScore, HPSv2\}, respectively.}
    \label{fig:mixing}
\end{figure}

\noindent\textbf{Interpolation.}
We demonstrate that, besides quality enhancements, TextCraftor can be weighted and interpolated with original text embeddings to control the generative strength. 
As in Fig.~\ref{fig:weighting}, with the increasing weights of enhanced text embeddings, the generated image gradually transforms into the reward-enhanced style. \textbf{Style Mixing.}
We also show that different reward-finetuned models can collaborate together to form style mixing, as in Fig.~\ref{fig:mixing}. 


\subsection{Ablation Analysis} \label{ablation}

\begin{figure}[]
    \centering
    \includegraphics[width=\linewidth]{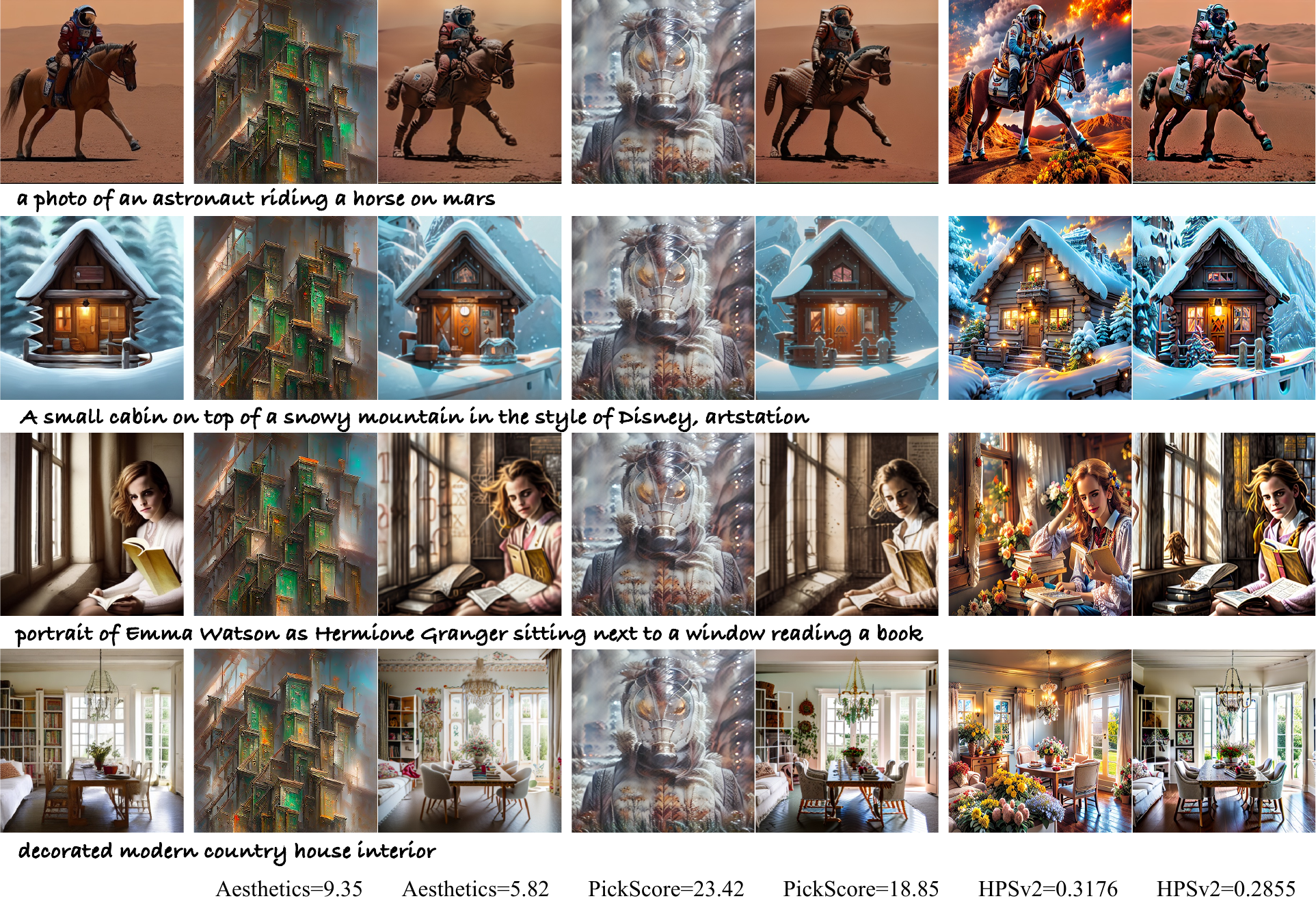}  
    \caption{\textbf{Ablation on reward models and the effect of CLIP constraint.} 
    The \emph{leftmost} column shows original images. Their averaged Aesthetics, PickScore, and HPSv2 scores are 5.49, 18.19, and 0.2672, respectively. 
    For the following columns, we show the synthesized images \emph{without} and \emph{with} CLIP constraint using different reward models. The reward scores are listed at the bottom.
   }
    \label{fig:ablate-clip-constraint}
\end{figure}

\noindent\textbf{Rewards and CLIP Constraint. }
We observe that simply relying on some reward functions might cause mode collapse problems. 
As in Fig.~\ref{fig:ablate-clip-constraint}, training solely on Aesthetics score or PickScore obtains exceptional rewards, but the model loses its generality and tends to generate a specific image that the reward model prefers. 
To conclude the root cause, not all reward models are pre-trained with large-scale fine-labeled data, thus lacking the capability to justify various prompts and scenarios. 
We see that HPSv2 shows better generality. 
Nevertheless, the CLIP constraint prevents the model from collapsing in all three reward regimes, while with reliable improvements in the corresponding scores. 

\noindent\textbf{Training and Testing Steps. }
As discussed in Section~\ref{sec:text-encoder-ft}, the reward fine-tuning introduces a long chain for gradient propagation. 
With the danger of gradient explosion and vanishing, it is not necessarily optimal to fine-tune over all timesteps. 
In Tab.~\ref{tab:ablation_steps}, we perform the analysis on the training and evaluation steps for \modelname. 
We find that training with $5$ gradient steps and evaluating the fine-tuned text encoder with $15$ out of the total $25$ steps gives the most balanced performance. 
In all of our experiments and reported results, we employ this configuration.

\begin{table}[h]
\small
\centering
\caption{Analysis of training and evaluation steps for fine-tuned text encoder. We report results on Parti-Prompts~\cite{yu2022scaling}. }
\scalebox{0.9}{
\begin{tabular}{ccccc}
\toprule
Train   & Test & Aes    & PickScore & HPSv2  \\
\hline
SDv1.5 & 25   & 5.2634 & 18.834    & 0.2703 \\
\hline
5       & 5    & 6.0688 & 19.195    & 0.2835 \\
5       & 10   & 6.3871 & 19.336    & 0.2847 \\
5       & 15   & 6.5295 & 19.355    & 0.2828 \\
5       & 25   & 6.5758 & 19.071    & 0.2722 \\
10      & 10   & 5.8680 & 19.158    & 0.2799 \\
15      & 15   & 5.3533 & 18.919    & 0.2735 \\
\bottomrule
\end{tabular}
}
\label{tab:ablation_steps}
\end{table}

We include more ablation studies on denoising scheduler and steps and reward weights in the supplementary material. 

\begin{figure*}[h]
    \centering
    \includegraphics[width=1\linewidth]{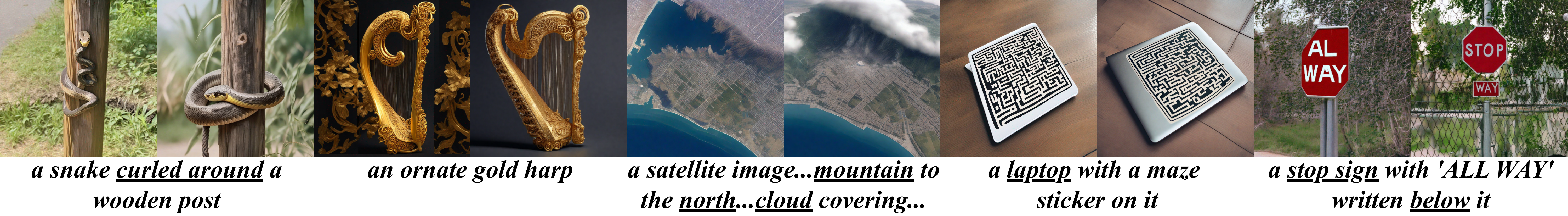} 
    \caption{Applying the fine-tuned SDv1.5 text encoder (ViT-L) under TextCraftor to SDXL can improve the generation quality, \eg, better text-image alignment. For each pair of images, the left one is generated using SDXL and the right one is from SDXL+TextCraftor. }
    \label{fig:app_sdxl}
\end{figure*}

\subsection{Discussion on Training Cost and Data}
TextCraftor is trained on $64$ NVIDIA A100 80G GPUs, with batch size $4$ per GPU. 
We report all empirical results of TextCraftor by training $10$K iterations, and the UNet fine-tuning (TextCraftor+UNet) with another $10$K iterations. 
Consequently, TextCraftor observes $2.56$ million data samples.
TextCraftor overcomes the collapsing issue thus eliminating the need for tricks like early stopping. 
The estimated GPU hour for TextCraftor is about $2300$. 
Fine-tuning larger diffusion models can lead to increased training costs. However, TextCraftor has a strong generalization capability.
As in Fig~\ref{fig:app_sdxl}, the fine-tuned SDv1.5 text encoder (ViT-L) can be directly applied to SDXL~\cite{podell2023sdxl} to generate better results (for each pair, \emph{left}: SDXL, \emph{right}: SDXL + TextCraftor-ViT-L). 
Note that SDXL employs two text encoders and we only replace the ViT-L one. 
Therefore, to reduce the training cost on larger diffusion models, one interesting future direction is to fine-tune their text encoder within a smaller diffusion pipeline, and then do inference directly with the larger model. 

\begin{figure}[h]
    \centering
    \includegraphics[width=1\linewidth]{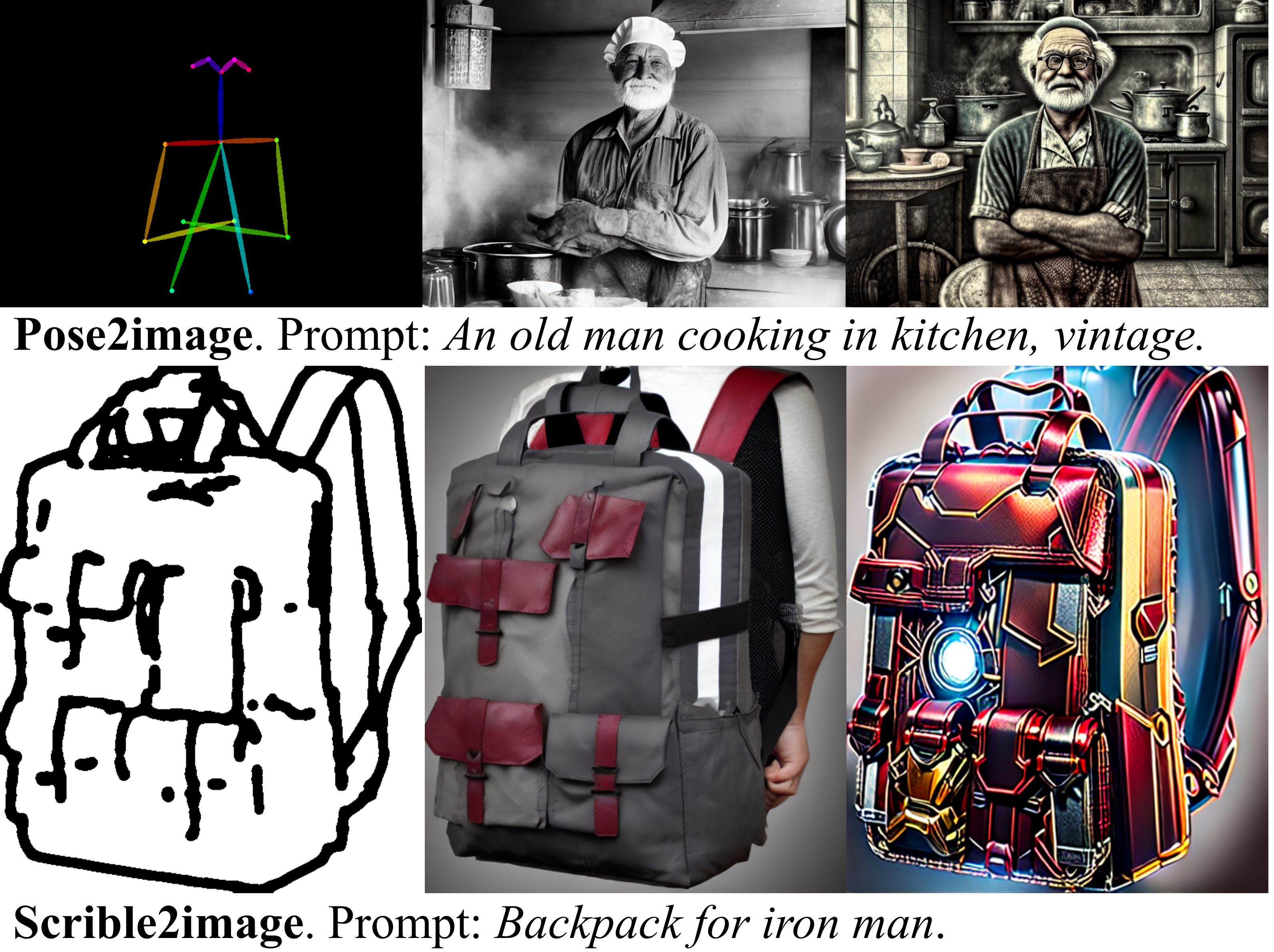}  
    \caption{Applying the fine-tuned SDv1.5 text encoder (ViT-L) under TextCraftor to ControlNet improves the generation quality. From left to right: input condition, SDv1.5, TextCraftor + SDv1.5 UNet. }
    \label{fig:app_controlnet}
\end{figure}

\begin{figure}[h]
    \centering
    \includegraphics[width=\linewidth]{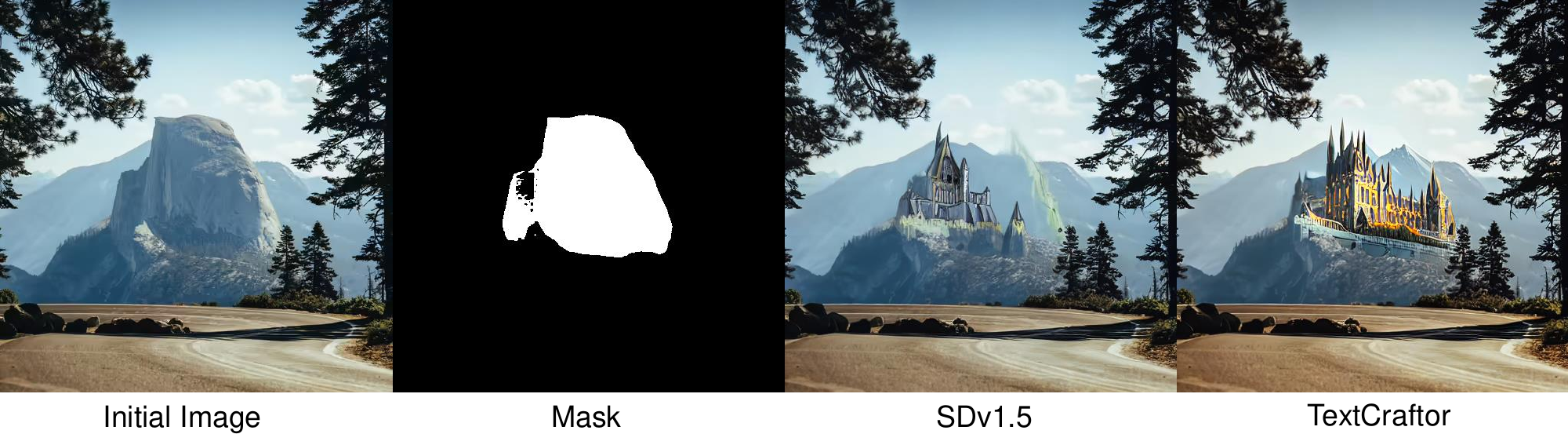}  
    \caption{Applying TextCraftor on inpainting task can improve the generation quality. The prompt in the example is \emph{concept art digital painting of an elven castle, inspired by lord of the rings}. }
    \label{fig:inpainting}
\end{figure}

\subsection{Applications}
We apply TextCraftor on ControlNet~\cite{zhang2023adding} and image inpainting for zero-shot evaluation (\ie, the pre-trained text encoder from TextCraftor is \emph{directly} applied on these tasks), as in Fig.~\ref{fig:app_controlnet} and Fig.~\ref{fig:inpainting}. 
We can see that TextCraftor can readily generalize to downstream tasks (with the same pre-trained baseline model, \ie, SDv1.5), and achieves better generative quality.

\section{Conclusion}\label{sec:conclusion}

In this work, we propose \modelname, a stable and powerful framework to fine-tune the pre-trained text encoder to improve the text-to-image generation. 
With only prompt dataset and pre-defined reward functions, \modelname~can significantly enhance the generative quality compared to the pre-trained text-to-image models, reinforcement learning-based approach, and prompt engineering.
To stabilize the reward fine-tuning process and avoid mode collapse, we introduce a novel similarity-constrained paradigm.
We demonstrate the superior advantages of \modelname~in different datasets, automatic metrics, and human evaluation.
Moreover, we can fine-tune the UNet model in our reward pipeline to further improve synthesized images. Given the superiority of our approach, an interesting future direction is to explore encoding the style from reward functions into specific tokens of the text encoder.



\clearpage
{
    \small
    \bibliographystyle{ieeenat_fullname}
    \bibliography{main}
}

\clearpage

\appendix


\section{Ablation on Scheduler and Denoising Steps}
The main paper uses a 25-step DDIM scheduler as the default configuration.
We provide an additional study on the choice of scheduler type and denoising steps in Tab.~\ref{tab:scheduler_steps}. 
We find that the widely adopted DDIM scheduler already yields satisfactory performance, which is comparable to or even better than second-order counterparts such as DPM. 
We also find that 25 denoising steps are good enough for generative quality, while increasing the inference steps to 50 has minimal impact on performance. 

\begin{table}[h]
\small
\centering
\caption{Ablation study on denoising scheduler and steps. We choose TextCraftor on all rewards as the baseline model. }
\label{tab:scheduler_steps}
\begin{tabular}{ccccc}
\toprule
Scheduler & Steps & Aesthetic & PickScore & HPSv2  \\
\hline
DDIM      & 25    & 5.8800    & 19.157    & 0.2805 \\
DDIM      & 50    & 6.0178    & 19.121    & 0.2769 \\
PNDM      & 25    & 5.0991    & 18.479    & 0.2632 \\
PNDM      & 50    & 5.9355    & 19.026    & 0.2748 \\
DPM       & 25    & 5.8564    & 19.145    & 0.2803 \\
Euler     & 25    & 5.9098    & 19.151    & 0.2804 \\
\bottomrule
\end{tabular}
\end{table}

\begin{table*}[h]
\small
\centering
\caption{Ablation study on different reward weights. The reported results are TextCraftor for text encoder only. }
\label{tab:ablate_reward_weight}
\begin{tabular}{cccc|cccc}
\toprule
\multicolumn{4}{c}{Weight}           & \multicolumn{4}{|c}{Score}               \\
\hline
CLIP & Aesthetic & PickScore & HPSv2 & CLIP   & Aesthetic & PickScore & HPSv2  \\
\hline
200  & 3         & 1         & 100   & 0.2952 & 6.0900    & 19.123    & 0.2757 \\
100  & 6         & 1         & 100   & 0.2385 & 7.1680    & 19.435    & 0.2730 \\
100  & 3         & 2         & 100   & 0.2615 & 6.6831    & 19.494    & 0.2798 \\
100  & 3         & 1         & 200   & 0.2711 & 6.4020    & 19.306    & 0.2850 \\
\bottomrule
\end{tabular}
\end{table*}

\section{Weight of Reward Functions}

With TextCraftor, it is free to choose different reward functions and different weights as the optimization objective. 
For simplicity, in the main paper, we scale all the rewards to the same magnitude. 
We report empirical results by setting the weight of CLIP constraint to 100, Aesthetic reward as 1, PickScore as 1, and HPSv2 as 100. 
In Tab.~\ref{tab:ablate_reward_weight}, we provide an additional ablation study on different reward weights. 
Specifically, we train TextCraftor with emphasis on CLIP regularization, Aesthetic score, PickScore, and HPSv2 respectively. 
We can observe that assigning a higher weight to a specific reward simply results in better scores. 
TextCraftor is flexible and readily applicable to different user scenarios and preferences. 
We observe the issue of repeated objects in Fig.~\ref{fig:more_visuals}, which is introduced along with UNet fine-tuning. 
\begin{figure}[h]
    \centering
    \includegraphics[width=1\linewidth]{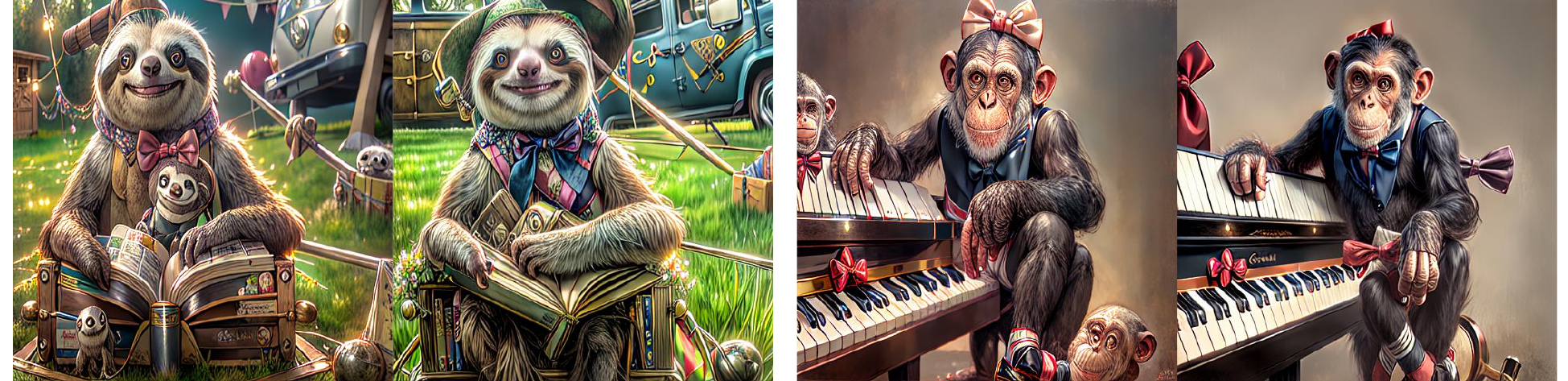}  
    \caption{Adjusting reward weights can further reduce artifacts (repeated objects.) }
    \label{fig:repeat_hps}
\end{figure}
Thus, we ablate TextCraftor+UNet fine-tuning with different weights of rewards. 
We find that HPSv2 is the major source of repeated objects.
We show in Fig.~\ref{fig:repeat_hps} that we can remove the repeated \emph{sloth} and \emph{chimpanzee} by using a smaller weight of HPSv2 reward.

\section{Interpretability}
\begin{figure*}[]
    \centering
    \includegraphics[width=\linewidth]{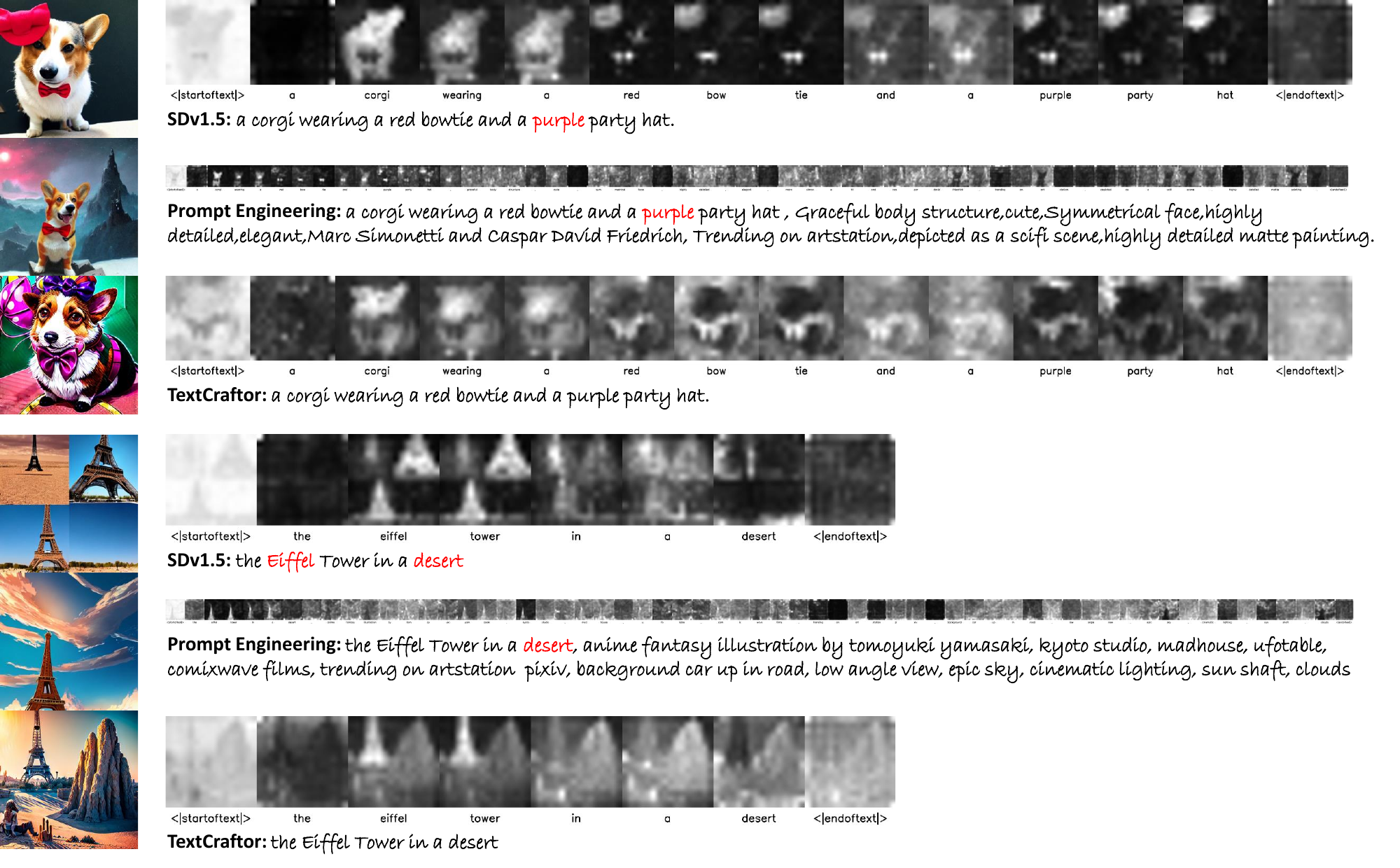}  
    \caption{Illustration of the enhanced semantic understanding in TextCraftor, visualized by the cross-attention heatmap. }
    \label{fig:interpret}
\end{figure*}

We further demonstrate the enhanced semantic understanding ability of TextCraftor in Fig.~\ref{fig:interpret}. 
Similar to Prompt to Prompt \cite{hertz2022prompt}, we visualize the cross-attention heatmap which determines the spatial layout and geometry of the generated image. 
We discuss two failure cases of the baseline model in Fig.~\ref{fig:interpret}. 
The first is misaligned semantics, as the \emph{purple} hat of the corgi. 
We can see that the hat in pixel space correctly attends to the word \emph{purple}, but in fact, the color is wrong (red). 
Prompt engineering does not resolve this issue. 
While in TextCraftor, color \emph{purple} is correctly reflected in the image. 
The second failure case is missing elements. 
SDv1.5 sometimes fails to generate desired objects, i.e.,  \emph{Eiffel Tower} or \emph{desert}, where there is hardly any attention energy upon the corresponding words.  
Prompt engineering introduces many irrelevant features and styles, but can not address the issue of the missing \emph{desert}. 
While with TextCraftor, both \emph{Eiffel Tower} and \emph{desert} are correctly understood and reflected in the image. 
We show that
(i) Finetuning the text encoder improves its capability and has the potential to correct some inaccurate semantic understandings. 
(ii) Finetuning text encoder helps to emphasize the core object, reducing the possibility of missing core elements in the generated image, thus improving text-image alignment, as well as benchmark scores.

\section{More Qualitative Results}
\begin{figure*}[h]
    \centering
    \includegraphics[width=\linewidth]{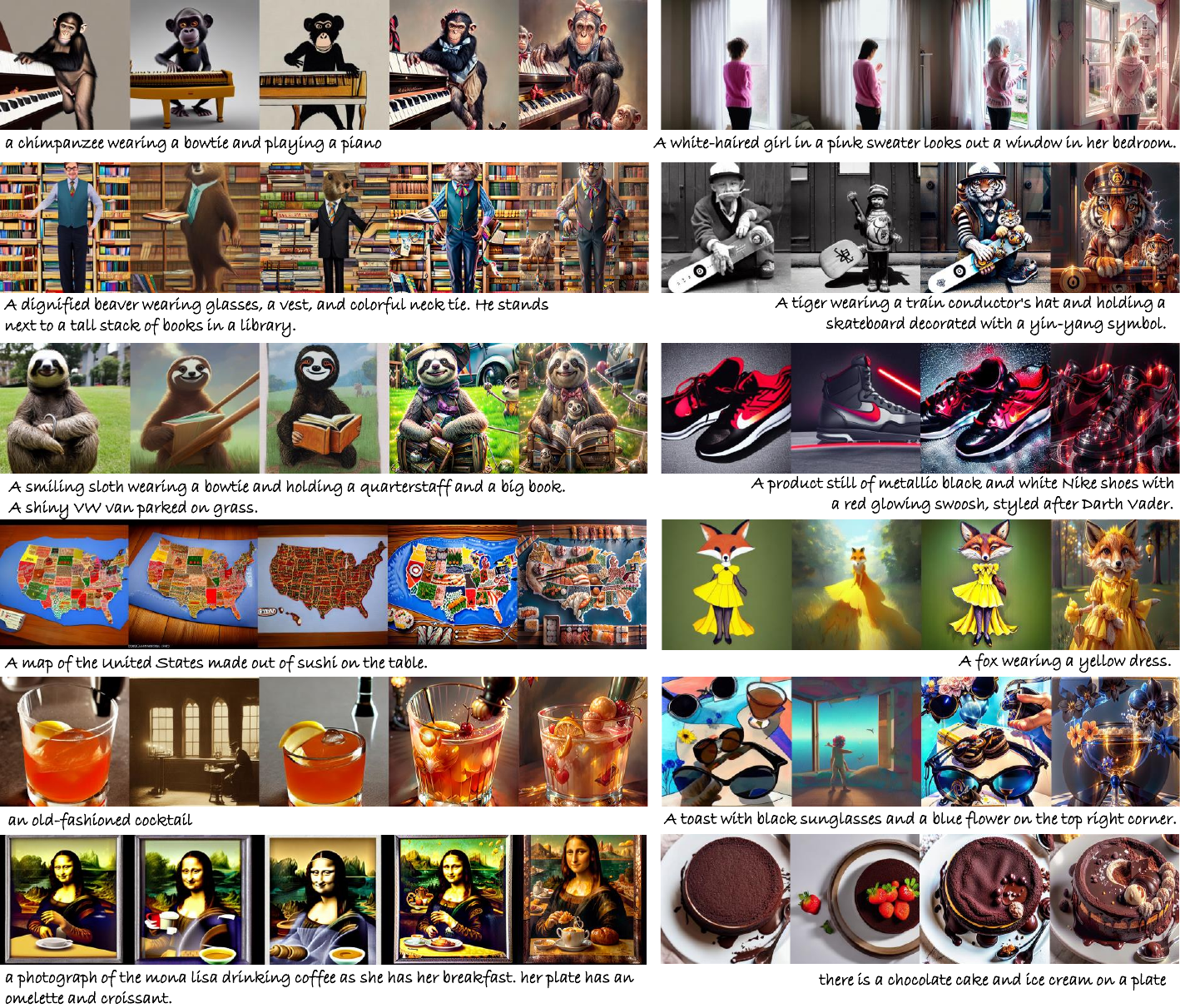}  
    \caption{\textbf{Additional visualizations.} \emph{\textbf{Left}}: generated images on Parti-Prompts, in the order of SDv1.5, prompt engineering, DDPO, \modelname, and \modelname~+ UNet. \emph{\textbf{Right}}: examples from HPSv2, ordered as SDv1.5, prompt engineering, \modelname, and \modelname~+ UNet. }
    \label{fig:more_visuals}
\end{figure*}

We provide more qualitative visualizations in Fig.~\ref{fig:more_visuals} to demonstrate the performance and generalization of TextCraftor. 

\end{document}